\documentclass[letterpaper]{article} 
\usepackage{aaai2026}  
\usepackage{times}  
\usepackage{helvet}  
\usepackage{courier}  
\usepackage[hyphens]{url}  
\usepackage{graphicx} 
\usepackage{amsfonts} 
\usepackage{natbib}  
\usepackage{caption} 
\usepackage{algorithm}
\usepackage{algorithmic}

\usepackage{newfloat}
\usepackage{listings}
\DeclareCaptionStyle{ruled}{labelfont=normalfont,labelsep=colon,strut=off} 
\floatstyle{ruled}
\newfloat{listing}{tb}{lst}{}
\floatname{listing}{Listing}
\usepackage{tikz}
\usetikzlibrary{positioning}
\usepackage{amsmath}
\usepackage{tabularray}
\title{Signed Dual Attention: Capturing Signed Dependencies in Time Series Forecasting}
\author {
    Balthazar Courvoisier\textsuperscript{\rm 1,\rm 2}\thanks{Corresponding author.},
    Tristan Cazenave\textsuperscript{\rm 2} 
}
\affiliations{
\textsuperscript{\rm 1}Queensfield AI Technologies, Paris, France\\
\textsuperscript{\rm 2}LAMSADE, Université Paris Dauphine - PSL, Paris, France\\

    balthazar.courvoisier@gmail.com
}

\newcommand\blfootnote[1]{%
  \begingroup
  \renewcommand\thefootnote{}\footnote{#1}%
  \addtocounter{footnote}{-1}%
  \endgroup
}

\begin{document}

\maketitle

\blfootnote{Accepted at the AI4TS Workshop, AAAI 2026. This is a non-archival preprint version.}

\begin{abstract}
Initially developed for natural language processing, Transformer architectures and attention mechanisms are now central to a wide range of deep learning models, including applications in time series forecasting. A standard attention mechanism, however, implicitly assumes homophilic interactions, limiting its ability to model data with positive and negative dependencies, such as time series. In this work, we introduce the Signed Dual Attention, a novel attention formulation that captures both positive and negative relational patterns without additional parameters. By leveraging a dual message-passing scheme inspired by correlation structures, Signed Dual Attention propagates both supportive and contrastive information within a single shared block, effectively achieving the expressiveness of two head attention without additional parameters. This module can be seamlessly integrated into existing architectures and can yield performance gains in certain situations, requiring signed relational modeling. This approach opens a pathway toward more expressive and parameter-efficient transformers.
\end{abstract}


\section{Introduction}

Transformers \cite{vaswani2023attentionneed} and attention mechanisms have become central to modern deep learning architectures, extending far beyond their initial applications in natural language processing. After impressive results in NLP tasks \cite{vaswani2023attentionneed}, attention-based models have demonstrated remarkable success in time series applications, such as energy consumption prediction and rate modeling. Architectures such as PatchTST \cite{nie2022time}, Chronos \cite{ansari2024chronoslearninglanguagetime}, and FEDformer \cite{zhou2022fedformerfrequencyenhanceddecomposed} heavily rely on attention mechanisms for temporal modeling. However, unlike NLP applications where large datasets are common, attention modules are sometimes applied in domains such as certain time series datasets, where available data is limited, making their large number of parameters a concern.

From a graph-theoretic perspective, the self-attention mechanism can be interpreted as message passing over a fully connected graph, where each token exchanges information with all others \cite{joshi2025transformersgraphneuralnetworks}. In this context, a single-head attention module assumes a homophilic structure, emphasizing similarity-based interactions between tokens and cannot model signed networks with negative relationships \cite{huang2019signedgraphattentionnetworks}. Real-world time series, for instance, often exhibit both positive and negative interactions : observations that are temporally close may display meaningful opposite trends and negative correlations \cite{10.1093/nar/gkp822, Agrawal_2019}. Standard attention mechanisms, which focus solely on positive scores after a softmax activation function, may thus fail to efficiently capture this structural behavior in temporal dependencies.

In this work, we propose a novel attention mechanism designed to capture both positive and negative relational patterns while maintaining a lightweight parameterization. Signed Dual Attention (SDA) introduces a dual message-passing formulation inspired by correlation structures: rather than using only high-valued attention scores, this mechanism explicitly leverages both strong positive and strong negative affinities. This formulation allows the model to propagate both supportive and contrastive information without duplicating parameters, effectively achieving the expressiveness of a two-head attention block within a single shared structure. 

\subsection{Contributions}

To the best of our knowledge, this is the first formulation of attention that integrates polarity-based message passing with explicit weight sharing outside the areas of graph neural networks (GNNs). While prior studies have explored forms of signed attention for GNNs \cite{huang2019signedgraphattentionnetworks, chen2023signgtsignedattentionbasedgraph, Grassia_2022}, this approach combines parameter efficiency with polarity sensitivity for the classic setting of a transformer, which means a fully connected graph. Moreover, we explicitly model this attention head as a two heads attention. Experiments on standard benchmarks for time series prediction demonstrate the potential benefits of SDA but call for further investigation.

In summary, the contributions are threefold:
\begin{enumerate}
\item We propose the Signed Dual Attention, a novel attention mechanism designed to jointly model signed relationships in graph-structured data.
\item We demonstrate that this module can be seamlessly integrated into existing architectures and provide performance gains for certain tasks.
\end{enumerate}
By rethinking the structure of attention through the lens of polarity and correlation, this work contributes to the development of more efficient and expressive transformer architectures for temporal modeling.

\section{Related Work}

\subsection{Attention Mechanisms in Deep Learning}

Attention mechanisms, first popularized in sequence-to-sequence models for natural language processing \cite{bahdanau2016neuralmachinetranslationjointly, vaswani2023attentionneed}, enable models to capture long-range dependencies more effectively than recurrent architectures, which are limited by sequential computation and vanishing gradients. 

Multi-head attention extends this idea by allowing the model to attend to multiple representation subspaces simultaneously, capturing distinct interaction patterns between tokens and improving expressiveness \cite{vaswani2023attentionneed}. This mechanism underpins the success of transformer architectures across diverse domains, including computer vision (e.g., ViT \cite{dosovitskiy2021imageworth16x16words}), video understanding \cite{bertasius2021spacetimeattentionneedvideo}, and time series forecasting \cite{zhou2022fedformerfrequencyenhanceddecomposed, lim2020temporalfusiontransformersinterpretable, ansari2024chronoslearninglanguagetime}. 

However, the quadratic complexity of attention with respect to sequence length imposes computational and memory constraints \cite{wang2021efficientconformerprobsparseattention}. To address this, numerous works propose sparse patterns, low-rank approximations, and memory-efficient architectures to preserve expressiveness while improving scalability \cite{zhou2021informerefficienttransformerlong, zhou2022fedformerfrequencyenhanceddecomposed}.

\subsection{Parameter-Efficient Attention}

While large-scale transformer models such as GPT-3 demonstrate the power of attention-based architectures \cite{brown2020languagemodelsfewshotlearners}, their vast parameter counts make them computationally expensive and prone to overfitting, particularly in low-data situations. To mitigate this, research has focused on parameter-efficient variants that reduce redundancy without compromising performance. Techniques such as low-rank factorization, parameter sharing, and sparse attention have achieved linear or sub-quadratic complexity while retaining expressive power \cite{wang2020linformerselfattentionlinearcomplexity, kitaev2020reformerefficienttransformer}. 

These methods enhance generalization and resource efficiency, enabling attention models to handle longer sequences and larger datasets. This approach draws on these principles to develop polarity-aware yet parameter-efficient attention mechanisms that scale effectively across diverse domains.

\subsection{Signed and Polarity-Aware Attention}

Standard transformers implicitly assume positive correlations between tokens, limiting their ability to capture negative or antagonistic relationships common in domains such as finance, biology, and energy systems \cite{10.1093/nar/gkp822, Agrawal_2019}. Modeling such interactions is essential for accurately representing inhibitory or contrastive dependencies.

In graph neural networks (GNNs), increasing attention has been paid to signed or contrastive relationships, where edges encode both positive and negative affinities. These settings, often characterized by heterophily, connections among dissimilar nodes, challenge conventional homophilous message passing \cite{Pan_2024, Agrawal_2019}. To handle this, signed message passing mechanisms explicitly differentiate between cooperative and antagonistic links. 

\citet{huang2019signedgraphattentionnetworks} introduced the Signed Graph Attention Network (SGAT), which maintains separate embedding spaces for positive and negative edges, enabling polarity-sensitive aggregation. Building on this, \citet{chen2023signgtsignedattentionbasedgraph} and \citet{Grassia_2022} refined polarity-aware attention to improve robustness and representation quality in relational domains. These studies show that incorporating edge polarity allows models to capture richer relational structures than unsigned attention.

Despite these advances, integrating polarity-sensitive message passing within fully connected transformer architectures remains underexplored. This work bridges this gap by extending polarity-aware mechanisms to transformers while preserving scalability and parameter efficiency.

\subsection{Autocorrelation Structure of Time Series}

In a univariate forecasting context, autocorrelation can be viewed as a form of attention, since it essentially highlights how current values are influenced by past values at various time lags \cite{wu2022autoformerdecompositiontransformersautocorrelation}. Autocorrelation in time series data has been extensively studied, particularly within the framework of classical statistical models such as ARIMA \cite{box1970distribution, hamilton1994time}. These models explicitly capture temporal dependencies, emphasizing how past observations influence future values. When representing temporal windows as nodes for token generation, an approach adopted by recent architectures such as PatchTST \cite{nie2022time}, the resulting networks cannot be straightforwardly categorized as either homophilic or heterophilic. This ambiguity arises because the partial autocorrelation structure of a time series can exhibit both positive and negative dependencies, depending on the underlying data-generating process \cite{10.1093/nar/gkp822, Agrawal_2019}. 

We posit that this duality is a defining characteristic that makes time series an especially suitable case for the proposed architecture. This approach implicitly assumes that each attention mechanism accounts simultaneously for both positive and negative relational effects. 

\section{Signed Dual Attention}

\subsection{SDA Block Formulation}

Our novel Signed Dual Attention (SDA) block extends the standard scaled dot-product attention to explicitly capture both positive and negative relational patterns without requiring new parameters. This approach is largely inspired by the behaviors that one can observe in the correlation matrix for time series, with positive and positive contribution coexists. This interpretation led us to consider that within a attention score matrix, both highly negative and highly positive scores must be selected for message propagation. However, the propagated message given a negative score should then be the opposite of the one for a positive score. Given the query, key, and value matrices $Q, K, V$, the SDA block computes :

\begin{small}
\begin{equation}
A^+ = \mathrm{softmax}\left(\frac{QK^\top}{\sqrt{d_k}}\right), \quad
A^- = \mathrm{softmax}\left(-\frac{QK^\top}{\sqrt{d_k}}\right)
\end{equation}
\end{small}
\begin{equation}
\mathrm{SDA}(Q,K,V) = (A^+ - A^-) V.
\end{equation}

Here, $A^+$ captures the supportive (homophilic) interactions, while $A^-$ encodes negative interactions. The subtraction ensures that both positive and negative relationships contribute to the aggregated representation and that the negative ones transmit opposite information compared to the positive ones.

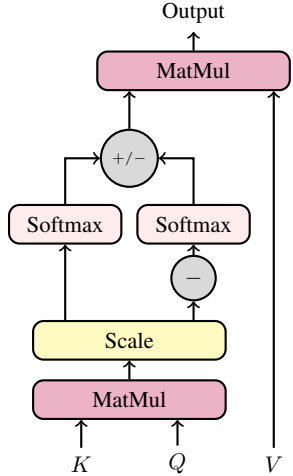
\begin{figure}[h]
\centering
\begin{tikzpicture}[every node/.style={scale=0.85}, thick]

\node[draw, fill=purple!30, rounded corners, minimum width=3cm, minimum height=0.6cm] (matmul1) {MatMul};
\node[draw, fill=yellow!30, rounded corners, minimum width=3cm, minimum height=0.6cm, above=0.25cm of matmul1] (scale) {Scale};
\node[draw, circle, fill=gray!30, minimum width=0.5cm, minimum height=0.5cm, above=0.25cm of scale, xshift=1cm] (minus) {$-$};
\node[draw, fill=pink!30, rounded corners, minimum width=1.75cm, minimum height=0.6cm, above=1cm of scale, xshift=-1cm] (softmax_1) {Softmax};
\node[draw, fill=pink!30, rounded corners, minimum width=1.75cm, minimum height=0.6cm, above=1cm of scale, xshift=1cm] (softmax_2) {Softmax};
\node[draw, circle, fill=gray!30, minimum width=0.5cm, minimum height=0.5cm, above=1.75cm of scale] (sum) {\tiny{$+/-$}};

\node[draw, fill=purple!30, rounded corners, minimum width=3cm, minimum height=0.6cm, above=0.5cm of sum, xshift=1cm] (matmul2) {MatMul};

\node[xshift=0.75cm, yshift=-1cm] (Q) {$Q$};
\node[xshift=-0.75cm, yshift=-1cm] (K) {$K$};
\node[xshift=2.25cm, yshift=-1cm] (V) {$V$};

\node[above=0.25cm of matmul2] (output) {Output};

\draw[->] (Q.north) -- (Q.north |- matmul1.south);
\draw[->] (K.north) -- (K.north |- matmul1.south);

\draw[->] (matmul1.north) -- (scale.south);
\draw[->] (scale.north -| minus.south) -- (minus.south);
\draw[->] (scale.north -| softmax_1.south) -- (softmax_1.south);
\draw[->] (V.north) -- (V.north |- matmul2.south);
\draw[->] (minus.north) -- (softmax_2.south);

\draw[->] (matmul2.north) -- (output.south);
\draw[->] (softmax_1.north) |- (sum.west);
\draw[->] (softmax_2.north) |- (sum.east);

\draw[->] (sum.north) -- (sum.north |- matmul2.south);

\end{tikzpicture}
\caption{Signed Dual Attention block.}
\label{fig:sda_block}
\end{figure}

The \noindent Figure~\ref{fig:sda_block} illustrates the conceptual flow of the SDA block. Queries and keys are combined via scaled dot-product to produce positive and negative attention matrices, which are then combined and multiplied by the value matrix to produce the final output. We perform the same number of matrix multiplications as in the standard attention layer, with the addition of a softmax operation, resulting in strong computational efficiency.

The SDA head can be seamlessly integrated within a multi-head architecture, analogous to the conventional attention module described by \citet{vaswani2023attentionneed}.

\subsection{Link between SDA and Two-Head Attention}

The Signed Dual Attention block can be interpreted as a constrained variant of a two-head self-attention mechanism. Consider a two-head attention layer with parameters:
\[
\begin{aligned}
(W_1^K, W_1^Q, W_1^V) &= (W^K, W^Q, W^V), \\
(W_2^K, W_2^Q, W_2^V) &= (-W^K,\, W^Q,\, -W^V).
\end{aligned}
\]

Let $H_1, H_2 \in \mathbb{R}^{T \times d}$ denote the output of each head. Following standard multi-head attention, their concatenation $H_{\text{cat}} \in \mathbb{R}^{T \times 2d}$ is projected by an output matrix $W^O$ to obtain the final representation $H$ \cite{vaswani2023attentionneed}:
\[
H = H_{\text{cat}} W^O.
\]

Choosing
\[
W^O =
\begin{bmatrix}
I_d \\[2pt]
I_d
\end{bmatrix}
\in \mathbb{R}^{2d \times d}
\]
yields a simple additive fusion, $H = H_1 + H_2$. Under this configuration, the output of the two-head mechanism exactly matches the SDA formulation in Eq.~(3), where one head encodes positive affinities and the other encodes negative affinities derived from the same similarity matrix.

Hence, SDA can be viewed as a \textit{parameter-tied} two-head attention, characterized by:
\begin{enumerate}
    \item Shared query projections across both heads,
    \item Negatively coupled key and value projections,
    \item An additive output projection replaces concatenation.
\end{enumerate}

This interpretation reveals that SDA effectively encodes both supportive and antagonistic interactions—analogous to dual-head attention—while using half the parameter count and maintaining the computational footprint of a single-head layer. The design thus emphasizes relational polarity without increasing model complexity.

\section{Experiments}

To evaluate the performance of the Attention block, we conducted experiments on 6 popular datasets. We benchmark the SDA block against the classic attention block of \cite{vaswani2023attentionneed} within two transformer-based models : Transformer \cite{vaswani2023attentionneed} and Informer \cite{zhou2021informerefficienttransformerlong}.

\subsubsection{Datasets.} We evaluate on several standard time series benchmarks, including ETT \cite{zhou2021informerefficienttransformerlong}, Electricity, Exchange \cite{lai2018modelinglongshorttermtemporal}, Traffic, and Weather \cite{zhou2022fedformerfrequencyenhanceddecomposed}. These datasets cover diverse domains such as energy consumption, exchange rates, traffic flow, and weather conditions.

\begin{table}[h!]
\centering
\caption{Number of distinct time steps in train and test sets.}
\label{tab:dataset_splits}
\begin{tabular}{lcc}
\hline
\textbf{Dataset} & \textbf{Train} & \textbf{Test} \\
\hline
Electricity & 18 293 & 5 237 \\
ETTm2 & 34 441 & 11 497 \\
ETTh2 & 8 521 & 2 857 \\
Exchange Rate & 5 192 & 1 494 \\
Traffic & 12 161 & 3 485 \\
Weather & 36 768 & 10 516\\
\hline
\end{tabular}
\end{table}

\subsubsection{Experimental Setup.} The experimental setup closely follows the one of \citet{zhou2022fedformerfrequencyenhanceddecomposed}. For each architecture, we compare the performance of the standard attention mechanism with that of SDA block, applied in both the encoder and decoder. We set $d_{\text{model}} = 512$ and $n_{\text{heads}} = 8$. Datasets are normalized following the procedure described in \cite{zhou2022fedformerfrequencyenhanceddecomposed}, where each time series is individually standardized using z-score normalization computed over the training split. Models are trained using the ADAM optimizer \cite{kingma2017adammethodstochasticoptimization} with a learning rate of $10^{-4}$ and a batch size of 32. An early stopping mechanism halts training if no improvement in validation loss is observed over three consecutive epochs. Performance is evaluated using mean squared error (MSE) and mean absolute error (MAE). Each experiment is repeated 3 times, and the reported results correspond to the mean values of the metrics. All deep learning models are implemented in PyTorch \cite{paszke2019pytorchimperativestylehighperformance}. We constrained ourselves in this work to long term univariate series forecasting with horizons of 24, 48 and 96 timestep and used an input length of 96 for all experiments. 

\subsubsection{Main Results.}

The results obtained with the proposed architecture are mixed. Overall, the integration of the module does not consistently enhance performance across the various datasets evaluated. Detailed results are presented in Tables~\ref{tab:transformer_results} and ~\ref{tab:informer_results}. The encouraging improvements are observed in the ETTm2 and ETTh2 datasets, while the proposed module performs notably worse on the Exchange dataset. 

\begin{table}
\centering
\caption{Influence on Transformer Architecture}
\label{tab:transformer_results}
\scalebox{0.76}{
\begin{tblr}{
  vline{3,6} = {1-2}{},
  vline{2-3,6} = {3-14}{},
  hline{2} = {3-8}{},
  hline{3,5,7,9,11,13} = {-}{},
}
         &     &       & SDA   &       &       & Classic &     \\
         &     & 24    & 48    & 96    & 24    & 48    & 96    \\
ECL      & MSE & 0.207 & 0.287 & 0.319 & \textbf{0.199} & \textbf{0.252} & \textbf{0.31}  \\
         & MAE & 0.337 & 0.398 & 0.42  & \textbf{0.328} & \textbf{0.37}  & \textbf{0.409} \\
Ettm2    & MSE & 0.024 & \textbf{0.058} & 0.137 & \textbf{0.02}  & 0.099 & \textbf{0.09}  \\
         & MAE & 0.112 & \textbf{0.173} & \textbf{0.187} & \textbf{0.102} & 0.246 & 0.234 \\
Etth2    & MSE & 0.103 & \textbf{0.149} & \textbf{0.231} & \textbf{0.101} & 0.159 & 0.238 \\
         & MAE & \textbf{0.25}  & \textbf{0.31}  & \textbf{0.387} & 0.252 & 0.318 & 0.394 \\
Exchange & MSE & 0.081 & 0.375 & 1.112 & \textbf{0.062} & \textbf{0.133} & \textbf{0.332} \\
         & MAE & 0.219 & 0.47  & 0.792 & \textbf{0.195} & \textbf{0.289} & \textbf{0.441} \\
Traffic  & MSE & 0.191 & 0.231 & \textbf{0.224} & \textbf{0.172} & \textbf{0.203} & 0.254 \\
         & MAE & 0.285 & 0.325 & \textbf{0.315} & \textbf{0.267} & \textbf{0.302} & 0.358 \\
Weather  & MSE & 0.003 & \textbf{0.01}  & 0.009 & \textbf{0.002} & 0.013 & \textbf{0.004} \\
         & MAE & 0.04  & 0.075 & 0.074 & \textbf{0.034} & \textbf{0.046} & \textbf{0.051}
\end{tblr}
}
\end{table}

\begin{table}
\centering
\caption{Influence on Informer Architecture}
\label{tab:informer_results}
\scalebox{0.76}{
\begin{tblr}{
  vline{3,6} = {1-2}{},
  vline{2-3,6} = {3-14}{},
  hline{2} = {3-8}{},
  hline{3,5,7,9,11,13} = {-}{},
}
~        & ~   &                & SDA   &                &                & Classic        &                \\
         &     & 24             & 48    & 96             & 24             & 48             & 96             \\
ECL      & MSE & 0.213          & 0.253 & 0.284          & \textbf{0.189} & \textbf{0.227} & \textbf{0.272} \\
         & MAE & 0.342          & 0.367 & 0.384          & \textbf{0.32}  & \textbf{0.347} & \textbf{0.374} \\
Ettm2    & MSE & \textbf{0.031} & 0.062 & 0.084          & 0.034          & \textbf{0.058} & \textbf{0.083} \\
         & MAE & \textbf{0.127} & 0.187 & \textbf{0.221} & 0.136          & \textbf{0.179} & \textbf{0.221} \\
Etth2    & MSE & 0.122          & 0.201 & 0.275          & \textbf{0.096} & \textbf{0.173} & \textbf{0.25}  \\
         & MAE & 0.274          & 0.359 & 0.423          & \textbf{0.24}  & \textbf{0.332} & \textbf{0.405} \\
Exchange & MSE & 0.092          & 0.179 & 0.421          & \textbf{0.071} & \textbf{0.145} & \textbf{0.367} \\
         & MAE & 0.243          & 0.335 & 0.524          & \textbf{0.212} & \textbf{0.309} & \textbf{0.48}  \\
Traffic  & MSE & 0.24           & 0.254 & 0.288          & \textbf{0.208} & \textbf{0.229} & \textbf{0.264} \\
         & MAE & 0.338          & 0.345 & 0.372          & \textbf{0.307} & \textbf{0.321} & \textbf{0.356} \\
Weather  & MSE & \textbf{0.003} & 0.009 & \textbf{0.005} & 0.004          & \textbf{0.007} & 0.006          \\
         & MAE & \textbf{0.042} & 0.062 & \textbf{0.052} & 0.044          & \textbf{0.054} & 0.053          
\end{tblr}
}
\end{table}

\subsubsection{Interpretation.} 
\begin{figure}[htbp]
    \centering
    \scalebox{0.6}{
        \includegraphics{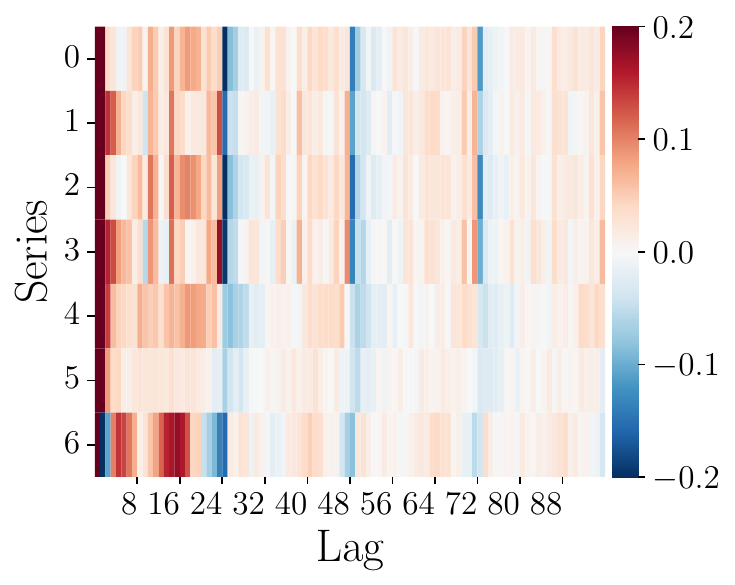}
    }
    \caption{Partial autocorrelation function (PACF) for each time series in the ETTh2 dataset. Each row corresponds to a single series, and columns represent lags (up to 96). Red/blue colors indicate positive/negative correlations with past values.}
    \label{fig:autocorr_etth2}
\end{figure}

\begin{figure}[htbp]
    \centering
    \scalebox{0.6}{
    \includegraphics{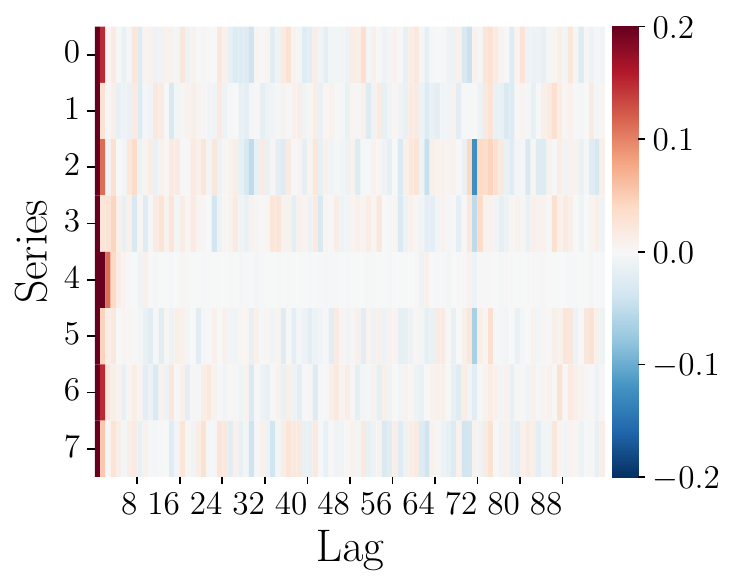}
    }
    \caption{Partial autocorrelation function (PACF) for each time series in the Exchange dataset. Each row corresponds to a single series, and columns represent lags (up to 96). Red/blue colors indicate positive/negative correlations with past values.}
    \label{fig:autocorr_exchange}
\end{figure}

To gain insight into these contrasting results, we examined the partial autocorrelation structures of the ETTh2 and Exchange datasets in figures ~\ref{fig:autocorr_etth2} and ~\ref{fig:autocorr_exchange} over a horizon of 96, corresponding to the one used in the forecasting experiments.

We observe that autocorrelations exhibit both positive and negative values quite prominently in the ETTh2 dataset, whereas the partial autocorrelation structure of the Exchange dataset is predominantly positive and concentrated on the first few lags. This SDA block assigns equal weighting to both positive and negative relationships. In scenarios where negative dependencies primarily represent noise, this symmetric treatment can potentially degrade performance. We hypothesize that the observed differences in autocorrelation structures between datasets partly explain the variation in the SDA block’s impact.

\section{Conclusion and Future Work}

Our proposed SDA block delivers contrasting results across different datasets. It appears more effective on datasets characterized by highly contrasted partial autocorrelation structures, where both positive and negative dependencies coexist.

These findings represent a promising first step. In future work, we plan to extend the evaluation to multivariate forecasting tasks and to integrate the SDA block within alternative architectures. Another important direction is the development of an encoder architecture specifically tailored to this attention mechanism, one that can preserve the signed aspects of relationships, thereby ensuring that temporal dependencies are faithfully captured throughout the model.

Finally, we see potential in learning adaptive weighting between the positive and negative attention components $A^{+}$ and $A^{-}$ instead of assigning them equal importance. This enhancement could improve performance in settings where negative interactions are weak or primarily noisy.

\section{Acknowledgments}

The authors gratefully acknowledge the support of the LAMSADE laboratory at PSL University for providing the computational resources necessary to conduct this research. The authors also thank Arnaud De Servigny, Pierre-Louis Barbarant and Samuel Bazaz for their valuable feedback and suggestions, which helped improve this work.

\small
\bibliography{aaai2026}

\end{document}